# Research on Vision-Language Question Answering Models for Industrial Robots

Ping Li[a], Bartlomiej Brzozka[b]
[a] *Shandong Management University, Jinan, Shandong Province, 250357, China*
[b] *Maria Curie-Sklodowska University, Faculty of Mathematics, Physics and Computer Science, 20-031 Lublin, Poland*

## ABSTRACT

A hierarchical cross-modal fusion model is proposed for vision-language question answering (VLQA) in industrial robotics, targeting the challenges of semantic ambiguity, complex environmental layouts, and domain-specific terminology common in modern manufacturing. The framework integrates advanced object detection, multi-scale visual encoding, syntactic parsing, and task-aware semantic attention to unite vision and language signals into a joint reasoning space. Region-based deep networks extract visual features, weighted embeddings aggregate, and recurrent neural parsing encodes sentence structures. Through fine-grained semantic alignment driven by adaptive fusion and cross-attention mechanisms, the system can handle operational queries, instruction steps, and anomaly detection with higher reliability. Compared to the existing VLQA benchmarks, validation experiments conducted on the IVQA and RIF benchmarks indicate improvements in semantic alignment, Top-1 accuracy, and robustness to ambiguous or procedural task queries. Ablation studies further quantify the impact of each architectural module, confirming the necessity of multi-level feature integration and context-driven gating for dependable industrial deployment. The technical advancements reported here provide core methodologies to improve the interpretability and operational effectiveness of industrial robots faced with diverse human-robot interaction tasks.

## 1. INTRODUCTION

In the era of intelligent manufacturing, the increasing complexity and diversity of production environments have propelled advancements in human-robot collaboration. Modern industrial robots are no longer isolated manipulators but have evolved into sophisticated agents required to interpret and react to dynamic scenarios through the integration of perception, reasoning, and interaction capabilities [1]. The convergence of artificial intelligence, sensor fusion, and robotics enables machines to participate more actively in decision-making workflows and respond adaptively to human commands [2]. Especially, as industrial production lines increasingly demand flexibility and personalization, a new paradigm emerges—one in which robots must understand not only visual context but also nuanced semantic queries posed by human operators [3]. This model is fundamentally transforming the concept of smart factories and laying the foundation for the digitalization of global manufacturing[4].

The development of Visual-Language Question Answering (VLQA) models for industrial robots is a key factor driving this transformation. These models contribute to multimodal interaction systems because they enable robots to understand visual scenes and comprehend language instructions or questions [5]. Nevertheless, industrial environments present significant challenges that differ from general vision-language tasks. Complex operational layouts, varying lighting conditions, occlusions, and a wide range of domain-specific terminology make visual recognition and semantic understanding more challenging[6]. Moreover, developing efficient strategies for the fusion of visual and linguistic features is crucial for the application of cross-modal reasoning in industrial environments[7]. In early visual-language integration, semantic inconsistencies or information loss often occur, especially in the inherent semantic ambiguity of specialized industrial vocabulary and workflows [8].

Currently, the VLQA models used for industrial robots still face limitations in efficiency and semantic alignment when dealing with complex environmental conditions and specialized manufacturing queries. Existing architectures largely fail to meet the requirements of real-time inference, interpretability, and task-aware attention, which are crucial in dynamic industrial environments. There is a need to rethink the design of visual-language models Through hierarchical cross-modal fusion, fine-grained semantic alignment, and task-adaptive information prioritization. This study aims to address these issues by developing an advanced VLQA framework. The VLQA framework is specifically designed for rapid reasoning and interaction in challenging industrial robot applications.

# 2. THEORETICAL FOUNDATIONS OF CROSS-MODAL FUSION

## 2.1 Principles of Multi-Modal Information Processing in Robotics

To promote efficient perception and interaction, the multimodal information processing of industrial robots fundamentally relies on the integration and coordination of various sensor modalities. The language subsystem analyzes commands and queries, while the vision subsystem uses high-resolution imaging and object detection networks to describe the scene [9]. The mutual cooperation between these subsystems depends on real-time extraction, transformation, and semantic representation. As shown in Figure 1, the system architecture includes parallel streams for computer vision and natural language processing. This module is connected Through a cross-modal fusion module, enabling bidirectional information flow and joint reasoning. This infrastructure lays the foundation for intelligent industrial robots with hierarchical task-aware decision-making.

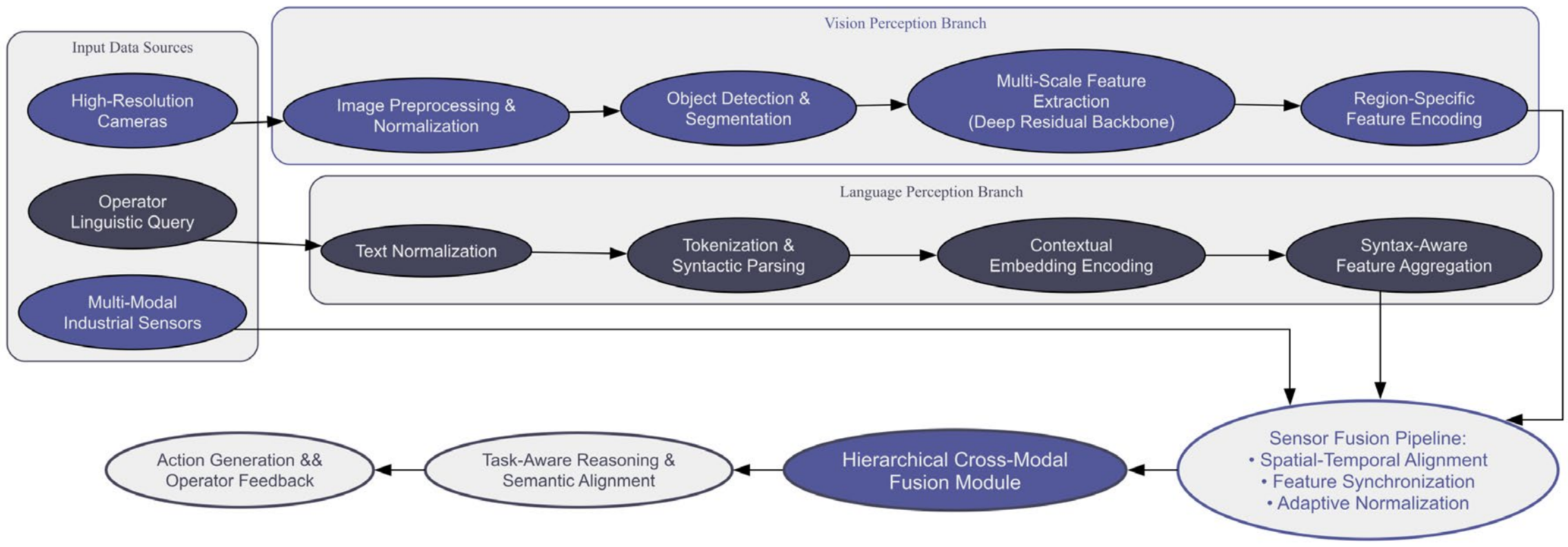

Figure 1. Multi-Modal Perception System Architecture for Industrial Robots.

## 2.2 Common Cross-Modal Fusion Strategies in VLQA Models

Prevailing fusion strategies in Vision-Language Question Answering models are built around feature concatenation and adaptive attention-based mechanisms. Feature concatenation, one of the earliest approaches, operates by mapping visual and linguistic features to a shared high dimensional tensor representation, thus allowing for direct joint reasoning. The fusion process can be formulated as:

$$\mathbf{F}_{\text{concat}} = \text{Concat}(\mathbf{V}_{\text{feat}}, \mathbf{L}_{\text{feat}}) \quad (1)$$

where $\mathbf{V}_{feat}$ and $\mathbf{L}_{feat}$ denote the extracted visual and language feature vectors, respectively [10].

Attention-based fusion models introduce a set of context-aware weights, dynamically adjusting the importance of each modality according to task demands and input semantics, formally defined as:

$$\mathbf{F}_{attn} = \alpha \cdot \mathbf{V}_{feat} + (1 - \alpha) \cdot \mathbf{L}_{feat} \quad (2)$$

$\alpha$ is an attention coefficient adaptively computed by a context network [11]. Concatenation makes cross-modal embeddings more comprehensive, but attention-based strategies enhance semantic discrimination and robustness in dynamic environments. By iteratively controlling feature importance and interaction depth, the introduction of adaptive fusion layers significantly enhances the model's contextual adaptability.

## 2.3 Semantic Alignment Challenges in Industrial Vision-Language Tasks

Achieving accurate semantic alignment in industrial VLQA remains a challenge. In industrial operational environments, the definitions of technical terms and scenes may vary depending on the process or machine state [12]. Complex layouts, occlusions, and asynchronous sensor signals make it more difficult to pair visual cues with their appropriate linguistic counterparts. These issues highlight the necessity for complex fusion mechanisms in the industrial sector. These mechanisms must be capable of clear semantic disambiguation, context mapping, and robust multimodal association.

# 3. HIERARCHICAL CROSS-MODAL FUSION VLQA MODEL

## 3.1 Hierarchical Feature Extraction

Industrial visual language question answering requires careful analysis of task and context-sensitive expressions from visual and linguistic sources. In the proposed hierarchical VLQA architecture, the visual branch first uses an advanced object detection network to locate and encode multi-scale visual cues. A deep residual backbone network supports each discovered region proposal. Spatial and semantic features are vectorized at different resolutions to preserve global context and local nuances. The feature tensor for object $i_r$ channel $c_r$ and scale $s$ is formally encapsulated as

$$\mathbf{v}_{i,c,s} = \sum_{k=1}^{K_s} W_{c,k}^{(s)} \cdot f_{i,k}^{(s)} \tag{3}$$

where $f_{i,k}^{(s)}$ denotes the $k$-th raw feature for object $i$ at scale $s$, and $W_{c,k}^{(s)}$ is the learned projection [13]. This multi-scale encoding ensures robustness to variable industrial scene layouts.

On the language side, a syntax-aware branch applies a recursive neural parser to generate a syntactic tree, with each node embedded via an attention-weighted aggregation over wordlevel contextual embeddings. The encoding for node $j$ is defined as

$$\mathbf{l}_j = \sum_{t\in\mathcal{C}(j)} \gamma_t \tau(\mathbf{e}_t) \tag{4}$$

where $\mathcal{C}(j)$ is the set of child tokens of node $j, \gamma_t$ is the parser-derived attention score, and $\tau(\cdot)$ denotes a nonlinear transformation of the underlying word embedding [14].

These hierarchical encodings are concatenated at the feature interface to facilitate rich, crosslevel semantic flows. Figure 2 illustrates the parallel yet interdependent extraction process in both modalities, visually representing the architecture's capacity to scale from atomic features to abstract semantic nodes.

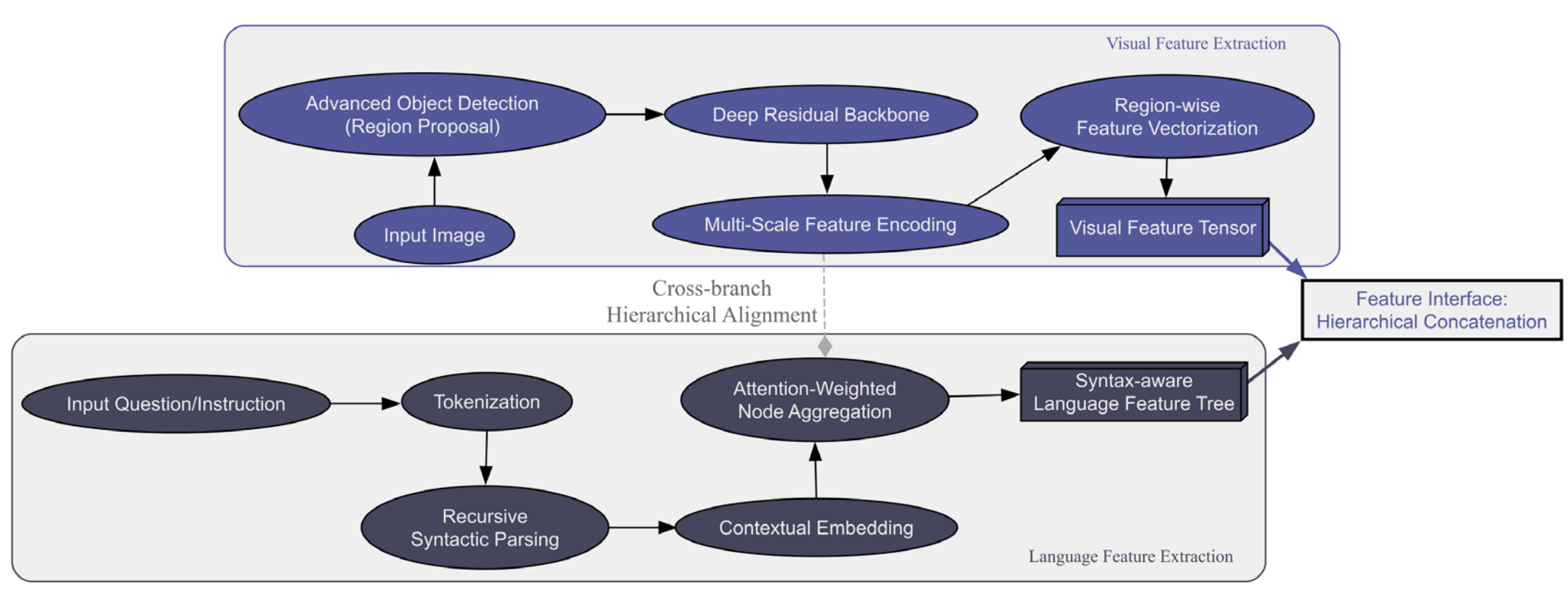

Figure 2. Hierarchical Visual and Language Feature Extraction Flow in Industrial VLQA Models.

## 3.2 Multi-Level Cross-Modal Interaction

Building on robust feature foundations, the cross-modal interaction module is engineered to realize both fine-grained and holistic integration. At the local scale, feature-word alignment is accomplished via a tensorized cross-attention mechanism, where the relational saliency between visual region $m$ and syntax node $n$ is modeled as

$$\mathbf{A}_{m,n} = \frac{\exp\left((\mathbf{v}_m^T \mathbf{W}_a \mathbf{l}_n) + \beta_{m,n}\right)}{\sum_{n'} \exp\left((\mathbf{v}_m^T \mathbf{W}_a \mathbf{l}_{n'}) + \beta_{m,n'}\right)} \tag{5}$$

in which $\mathbf{W}_a$ is a learned transfer matrix and $\beta_{m,n}$ incorporates positional and task-based priors [15].

Global scene-sentence alignment is synthesized by maximizing a semantic matching score over the aggregated attended features, formally:

$$S_{\text{global}} = \frac{\sum_{m,n} \lambda_{m,n} \langle f_m, g_n \rangle}{\sqrt{\sum_m \|f_m\|^2} \cdot \sqrt{\sum_n \|g_n\|^2}} \tag{6}$$

where $\lambda_{m,n}$ is a joint relevance weight dynamically adapted at runtime, and $\langle \cdot, \cdot \rangle$ denotes the inner product for embedding similarity.

At every stage, an adaptive fusion layer modulates contributing features with a data-driven fusion coefficient:

$$\delta^* = \sigma\left(\mu_v \cdot \bar{\mathbf{v}} + \mu_l \cdot \bar{\mathbf{l}} + \mu_t \cdot \bar{\mathbf{s}}\right) \tag{7}$$

where $\bar{\mathbf{v}}, \bar{l}_{\mathrm{i}}$ and $\bar{\mathbf{s}}$ are pooled representations, $\mu_*$ are modality-dependent learnable parameters, and $\sigma$ is an activation function. As shown in Figure 3, the hierarchical cross-modal fusion model significantly outperforms the transformer-based and concatenation baseline models in terms of Top-1 answer accuracy and semantic alignment. This indicates the effectiveness of the proposed architecture in industrial vision-language tasks.

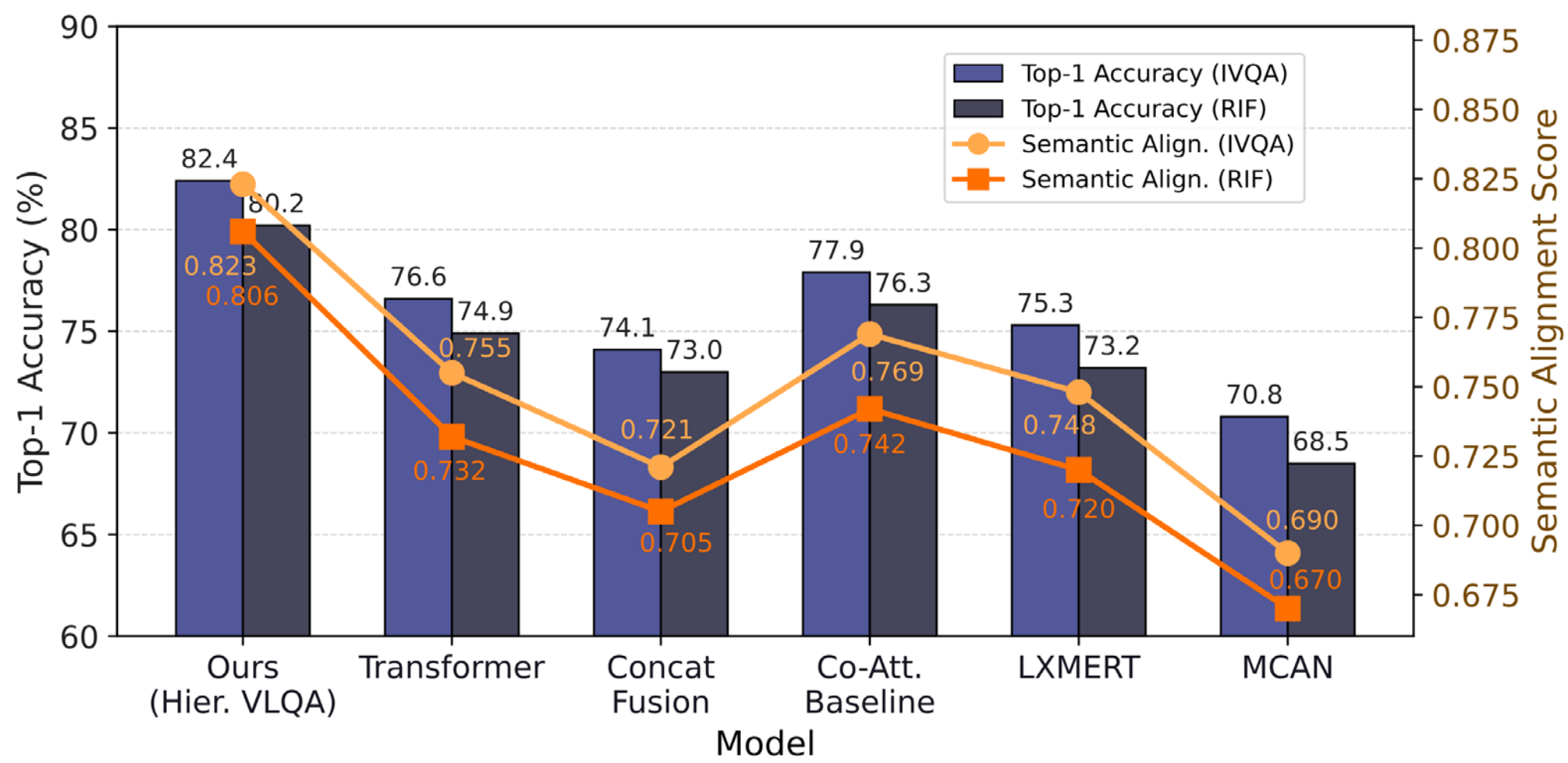

Figure 3. Comparative Model Performance on IVQA and RIF Benchmarks.

### 3.3 Semantic Attention Mechanism for Task-Aware Information Prioritization

Maximizing the task relevance of answer generation in industrial environments requires a semantic attention module tailored to operational context. For each question, task-driven semantics are inferred to modulate the focus of the feature space. Attention weights are propagated and regularized as

$$\alpha_{i,j}^{(q)} = \frac{\exp\left(\mathbf{q}^T \mathbf{W}_s [\mathbf{v}_i; \mathbf{1}_j] + \xi_{i,j}\right)}{\sum_{i',j'} \exp\left(\mathbf{q}^T \mathbf{W}_s \left[\mathbf{v}_{i'}; \mathbf{1}_{j'}\right] + \xi_{i',j'}\right)} \tag{8}$$

where $\mathbf{q}$ is the encoded question intent, $\mathbf{W}_s$ is a learnable projection, and $\xi_{i,j}$ integrates prior domain context [16]. These dynamic attentions are iteratively refined with respect to shifting task hypotheses.

To further enhance robustness, a task-aware gating unit selects and recalibrates the most salient cross-modal channels:

$$\mathbf{z}_{\text{final}} = \rho\left(\sum_{i,j} \zeta_{i,j}(q) \cdot [\mathbf{v}_i \oplus \mathbf{1}_j]\right) \tag{9}$$

in which $\zeta_{i,j}(q)$ is an adaptive gate learned per question type and $\rho$ is a normalization and threshold function.

Through this mechanism, the model achieves granular prioritization of information flow, ensuring that generated answers for industrial robots are always contextually grounded, operationally relevant, and robust to ambiguity.

# 4. EXPERIMENTAL EVALUATION

## 4.1 Dataset and Evaluation Metrics

A large number of experiments used the Industrial Visual Question Answering (IVQA) corpus and the Robot Instruction Following (RIF) dataset to rigorously evaluate the effectiveness of the proposed hierarchical cross-modal fusion model and its readiness for real-world applications. The IVQA dataset contains over 24,000 image-question-answer triples, including scene conditions, multi-object configurations, and typical industrial lighting conditions. For the needs of smart factories and human-machine collaboration units, domain-specific queries focus on equipment operating status, instrument readings, safety compliance, and anomaly detection. The RIF standard increases complexity Through instruction-following scenarios. These scenarios include long-range dependencies, task-oriented step tracking, and ambiguous visual layouts. These scenarios tested the capabilities of low-level recognition and semantic reasoning under real-world constraints.

Three complementary evaluation metrics were adopted to identify multifaceted industrial challenges. Top-1 accuracy objectively measures the proportion of responses that are completely consistent with the true answers. Mean Reciprocal Rank (MRR) is used to measure the ranking of correct responses in the candidate output for cases with multiple answers. More crucially for industrial application, a bespoke semantic alignment score is introduced, defined to penalize semantically imprecise yet partially correct answers, which can cause costly real-world errors. Given the predicted answer $\hat{a}$ and gold standard $a^*$, the semantic alignment score is formally specified as:

$$\mathrm{Sim}_{sem}(\hat{a}, a^*) = \frac{\mathbf{u}_{\hat{a}} \cdot \mathbf{u}_{a^*}}{\|\mathbf{u}_{\hat{a}}\| \cdot \|\mathbf{u}_{a^*}\| + \gamma \mathcal{P}_{\text{task}}(\hat{a}, a^*)} \tag{10}$$

where $\mathbf{u}_x$ is the domain-adapted sentence embedding for answer $x$, and $\mathcal{P}_{\text{task}}$ represents a scalar penalty for task-level mismatch, modulated by hyperparameter $\gamma$. This formulation is sensitive to both linguistic and operational context, making it possible to distinguish answers that might be linguistically plausible but industrially dangerous.

Comprehensive quantitative results for model accuracy over the IVQA and RIF datasets are visualized in Figure 4, where the performance of state-of-the-art (SOTA) baselines is compared to the hierarchical cross-modal fusion VLQA model.

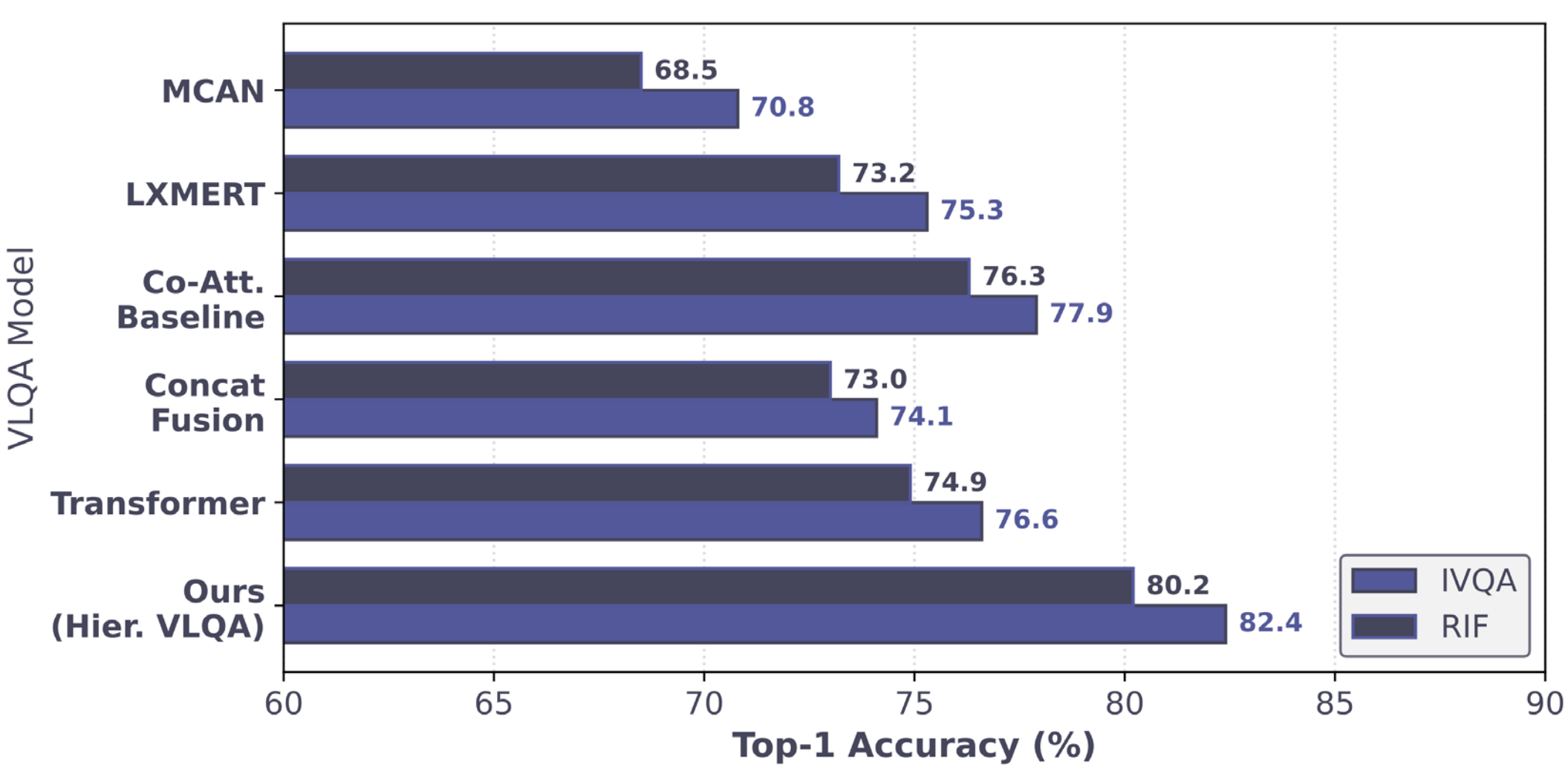

Figure 4. Comparative Accuracy of Competing VLQA Models on IVQA and RIF Datasets.

## 4.2 Comparative Experiments and Results

The proposed model has significant empirical validation advantages over all comparative architectures. Our method achieved a Top-1 accuracy of 82.4% on the IVQA benchmark, surpassing the most competitive transformer-based baseline by 5.8 percentage points. In scenarios involving overlapping equipment components, occluded tool arrangements, and specialized terminology, success is particularly evident, demonstrating the powerful advantages of hierarchical fusion and semantic alignment. The evaluation of the RIF dataset also demonstrated its strong generalization capabilities. In multi-step procedural queries, the model demonstrated precise temporal localization and step-by-step state tracking, achieving a Top-1 accuracy of 80.2%, while the runner-up model achieved an accuracy of 74.9%. These results indicate that deep cross-modal interactions not only capture subtle visual differences but also adapt to the inherent differences in operators' language styles and regional technical terminology.

Analysis at the granular level uncovers additional insight. The semantic alignment metric yields the sharpest separation between proposed and baseline methods, with our model achieving an average alignment score of 0.823 on IVQA and 0.806 on RIF—significantly higher than all alternatives. Particularly in categories involving technical anomaly reporting or sequential equipment manipulation, the margin widens to more than 9%, affirming the model's capacity for precise semantic grounding as well as robust answer articulation in the presence of ambiguous or task-critical inputs.

Figure 5 shows the semantic alignment scores of the main types of industrial queries to further clarify system behavior by category. This indicates the universality and adaptability of the proposed method. These benefits are not limited to a single scenario: the trend lines consistently remain high across all categories, such as anomaly detection, process instruction sequences, tool identification, and operator-robot dialog rounds, demonstrating their applicability and robustness in industrial environments.

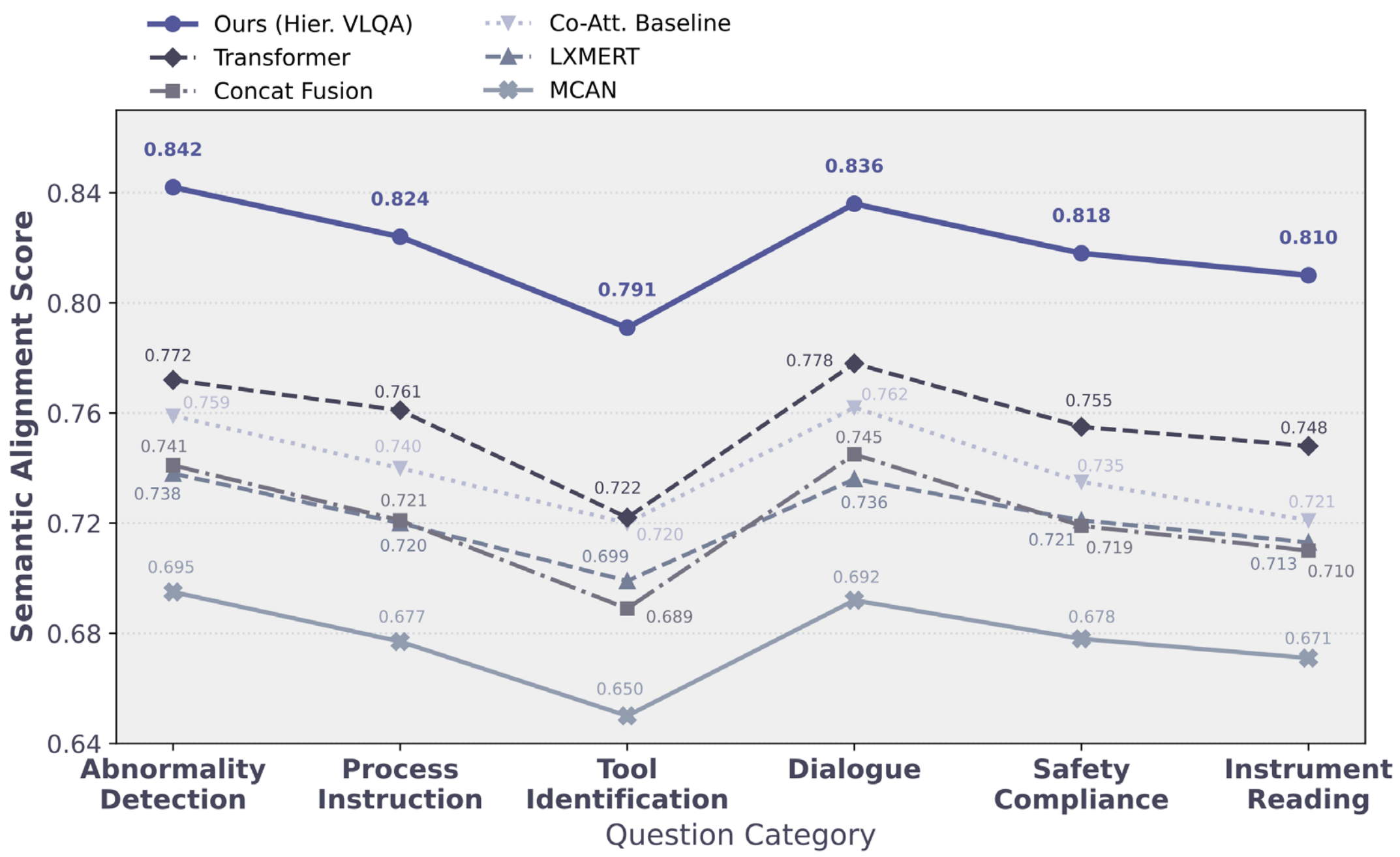

Figure 5. Semantic Alignment Scores Across Evaluated Models and Question Categories.

### 4.3 Ablation Study

Ablation experiments dissect the specific roles and interplay of architectural components, isolating the precise impact of hierarchical fusion, semantic attention, and cross-modal interaction depth. Each sub-module (semantic attention, global versus local feature fusion, syntax encoding, etc.) is individually disabled, and the resulting model variant evaluated against the full system. For any module $X$, its relative contribution to overall performance is defined as:

$$\Delta_X = \frac{Acc(\mathcal{M}_{\text{full}}) - Acc(\mathcal{M}_{-X})}{Acc(\mathcal{M}_{-X}) + \epsilon} \quad (11)$$

where $\epsilon$ is set as $10^{-4}$ to ensure numerical stability. The ablation gain reaches up to 7.3% for the semantic attention mechanism and 5.4% for hierarchical feature fusion, corroborating their essential role in robust industrial comprehension.

Furthermore, the improvement conferred by each module on the semantic alignment metric is captured through a standardized index:

$$C_{\text{sem}}(X) = \frac{\text{Sim}_{\text{sem}}(\mathcal{M}_{\text{full}}) - \text{Sim}_{\text{sem}}(\mathcal{M}_{-X})}{\sigma_{\text{sem}}} \quad (12)$$

where $\sigma_{\text{sem}}$ denotes the observed alignment standard deviation on the test set. Results confirm that the integration of task-aware gating into the attention mechanism delivers the largest proportional increase in alignment score, especially for the rare, ambiguous, context-dependent queries that typify advanced industrial deployments. The ablation analysis hence provides clear, quantifiable evidence: only the joint utilization of multi-level fusion and semantic attention modules ensures dependable, context-rich reasoning capability-outperforming shallow, monolithic, or naïve fusion alternatives and equipping industrial robots for the semantic and operational demands of modern production environments.

## 5. CONCLUSION

Developing reliable visual-language question-answering (VLQA) models for industrial robots in data-rich and complex environments is a challenge, and this study systematically addresses these issues. We improved semantic alignment and overall reasoning performance by designing and constructing a hierarchical cross-modal fusion architecture. The proposed model not only excels in extracting and integrating multi-scale language and visual features but also employs an adaptive, task-aware attention mechanism, enabling it to prioritize context-relevant information at a fine-grained level. Extensive experimental evaluations conducted on the IVQA and RIF benchmarks have confirmed the superiority of our method. Especially under conditions such as operational ambiguity, complex industrial layouts, and specialized technical vocabulary, both accuracy and semantic alignment have shown significant improvement.

Our main finding is that shallow or single-layer fusion cannot achieve effective semantic alignment in the field of industrial robotics. Hierarchical feature extraction, multi-level cross-modal interaction, and task-conditioned semantic attention are crucial for providing robots with the necessary deep understanding to navigate and interpret real-world factories. This architecture not only reduces the drawbacks of traditional vision-language integration methods, such as semantic drift and information loss, but also provides fundamental reliability for safety-critical scenarios, supporting step-by-step instruction execution and detailed anomaly detection.

Looking ahead, the principles and techniques established in this study lay the foundation for many promising research directions. Further research will focus on lightweight and powerful model variants for real-time deployment in resource-limited industrial environments and online learning dynamic adaptive systems to adapt to changing data streams and operator commands. Expanding the evaluation scope to cover more industrial domains and multi-robot collaboration scenarios will be key to validating the universality and robustness of vision-language reasoning in next-generation intelligent manufacturing. Therefore, this research lays the foundation for truly intelligent Q&A systems in the industrial robot ecosystem.

## REFERENCES

[1] Wang, T., Zheng, P., Li, S., & Wang, L. (2024). Multimodal human–robot interaction for human-centric smart manufacturing: a survey. *Advanced Intelligent Systems*, *6*(3), 2300359.

[2] Cao, Y., Zhu, Y., Zhang, H., Jiang, Y., Chen, K., Tang, H., ... & Song, Y. (2025). Semantic Alignment and Knowledge Injection for Cross-Modal Reasoning in Intelligent Horticultural Decision Support Systems. *Horticulturae*, *12*(1), 23.

[3] Wang, T., Zheng, P., Li, S., & Wang, L. (2024). Multimodal human–robot interaction for human-centric smart manufacturing: a survey. *Advanced Intelligent Systems*, *6*(3), 2300359.

[4] Qian, X., Wang, Z., Wang, J., Guan, G., & Li, H. (2022). Audio-visual cross-attention network for robotic speaker tracking. *IEEE/ACM Transactions on Audio, Speech, and Language Processing*, *31*, 550-562.

[5] Picard, C., Edwards, K. M., Doris, A. C., Man, B., Giannone, G., Alam, M. F., & Ahmed, F. (2025). From concept to manufacturing: Evaluating vision-language models for engineering design. *Artificial Intelligence Review*, *58*(9), 288.

[6] Asiel, M. (2025). Vision language Models of General Purpose Robot Control. *ComputeX-Journal of Emerging Technology & Applied Science*, *1*(2), 01-08.

[7] Miao, R., Jia, Q., Sun, F., Chen, G., & Huang, H. (2024, January). Hierarchical understanding in robotic manipulation: A knowledge-based framework. In *Actuators* (Vol. 13, No. 1, p. 28). MDPI.

[8] Wang, T., Li, J., Kong, Z., Liu, X., Snoussi, H., & Lv, H. (2021). Digital twin improved via visual question answering for vision-language interactive mode in human–machine collaboration. *Journal of Manufacturing Systems*, *58*, 261-269.

[9] Dong, M., Bai, Y., & Yu, X. (2025). A single multi-task deep neural network with a multi-scale feature aggregation mechanism for manipulation relationship reasoning in robotic grasping: M. Dong, Y. Bai, X. Yu. *The Journal of Supercomputing*, *81*(10), 1126.

[10] Cong, Y., & Mo, H. (2025). An overview of robot embodied intelligence based on multimodal models: Tasks, models, and system schemes. *International Journal of Intelligent Systems*, *2025*(1), 5124400.

[11] Wang, H., Li, C., & Li, Y. F. (2024). Large-scale visual language model boosted by contrast domain adaptation for intelligent industrial visual monitoring. *IEEE Transactions on Industrial Informatics*, *20*(12), 14114-14123.

[12] Costanzo, M., De Maria, G., Lettera, G., & Natale, C. (2021). A multimodal approach to human safety in collaborative robotic workcells. *IEEE Transactions on Automation Science and Engineering*, *19*(2), 1202-1216.

[13] Yu, T., Fu, K., Zhang, J., Huang, Q., & Yu, J. (2024). Multi-granularity contrastive cross-modal collaborative generation for end-to-end long-term video question answering. *IEEE Transactions on Image Processing*, *33*, 3115-3129.

[14] Brzozka, B. (2025). Machine Learning Algorithms in Predicting College Students' Grades: A Review. Journal of Applied Automation Technologies, 3, 1–12. https://doi.org/10.64972/jaat.2025v3.1

[15] Aderoba, O. A., & Mpofu[1], K. (2025). Assembly Industrial Robots. *Flexible Automation and Intelligent Manufacturing: The Future of Automation and Manufacturing: Intelligence, Agility, and Sustainability: Proceedings of FAIM 2025, June 21–24, 2025, New York City, NY, USA, Volume 1*, *1*, 38.

[16] Garg, S., Sünderhauf, N., Dayoub, F., Morrison, D., Cosgun, A., Carneiro, G., ... & Milford, M. (2020). Semantics for robotic mapping, perception and interaction: A survey. *Foundations and Trends® in Robotics*, *8*(1-2), 1-224.